\documentclass[letterpaper]{article} 

\usepackage{aaai24}  
\nocopyright  

\usepackage{times}  
\usepackage{helvet}  
\usepackage{courier}  
\usepackage[hyphens]{url}  
\usepackage{graphicx} 
\usepackage{natbib}  
\usepackage{caption} 
\usepackage{algorithm}
\usepackage{algorithmic}

\newcommand{\ours}{\textbf{\textsc{CIRS}}}

\newcommand{\bizhen}[1]{\textcolor{black}{#1}}
\newcommand{\aaai}[1]{\textcolor{black}{#1}}

\usepackage{amsmath}
\usepackage{booktabs}
\usepackage{colortbl}
\usepackage{hhline}
\usepackage{multirow}

\usepackage{newfloat}
\usepackage{listings}
\DeclareCaptionStyle{ruled}{labelfont=normalfont,labelsep=colon,strut=off} 
\floatstyle{ruled}
\newfloat{listing}{tb}{lst}{}
\floatname{listing}{Listing}
\title{When Do Program-of-Thought Works for Reasoning?}

\author{
    Zhen Bi$^{\spadesuit \diamondsuit}$,
    Ningyu Zhang$^{\spadesuit \diamondsuit}$\thanks{Corresponding Author.},
    Yinuo Jiang$^{\spadesuit \diamondsuit}$,
    Shumin Deng$^{\clubsuit}$,
    \\
    Guozhou Zheng$^{\spadesuit \diamondsuit \heartsuit}$,
    Huajun Chen$^{\spadesuit \diamondsuit\heartsuit}$\footnotemark[1],
}

\affiliations{
    
    $^\spadesuit$Zhejiang University 
    $^\diamondsuit$Zhejiang University - Ant Group Joint Laboratory of Knowledge Graph
    \\
    $^\heartsuit$Donghai Laboratory
    $^\clubsuit$NUS-NCS Joint Lab, National University of Singapore
    \\
    \{bizhen\_zju, zhangningyu, 3200100732, guozhou, huajunsir\}@zju.edu.cn, 
    shumin@nus.edu.sg \\

}

\begin{document}

\maketitle

\begin{abstract}

\aaai{In the realm of embodied artificial intelligence, the reasoning capabilities of Large Language Models (LLMs) play a pivotal role.}
Although there are effective methods like \textit{program-of-thought}  prompting for LLMs which uses programming language to tackle complex reasoning tasks, the specific impact of code data on the improvement of reasoning capabilities remains under-explored.
To address this gap, we propose complexity-impacted reasoning score (\ours), which combines structural and logical attributes, to measure the correlation between code and reasoning abilities. Specifically, we use the abstract syntax tree to encode the structural information and calculate logical complexity by considering the difficulty and the cyclomatic complexity. Through an empirical analysis, we find not all code data of complexity can be learned or understood by LLMs. Optimal level of complexity is critical to the improvement of reasoning abilities by program-aided prompting. Then we design an auto-synthesizing and stratifying algorithm, and apply it to instruction generation for mathematical reasoning  and code data filtering for code generation tasks. Extensive results demonstrates the effectiveness of our proposed approach. 
Code will be integrated into the EasyInstruct framework\footnote{\url{https://github.com/zjunlp/EasyInstruct}}.
\end{abstract}

\section{Introduction}

Large language models (LLMs) \cite{openai2023gpt4, anil2023palm}, have emerged as a general-purpose problem-solving methodology for embodied artificial intelligence.
In the realm of embodied AI, the reasoning capabilities of LLMs play a pivotal role, especially when agents need to comprehend the semantic intricacies of their environment for effective control \cite{DBLP:conf/www/ChenZXDYTHSC22,DBLP:conf/corl/HuangXXCLFZTMCS22, DBLP:journals/corr/abs-2307-05973, DBLP:journals/corr/abs-2305-16291}. 
Recent approaches \cite{POT, PAL, binder}, which we term \textit{program-of-thought}, leverages programming language as a superior prompting mechanism for complex reasoning tasks.
In contrast to \textit{chain-of-thought}  prompting \cite{cot}, \textit{program-of-thought} prompting disentangles the problems into executable code segments and address them step-by-step.
However, the correlation between the programming language utilzation and the improvement in reasoning ability for LLMs is under-studied.
The essential question still remains: \textbf{When do  program-of-thought prompting works for reasoning\footnote{In this work, we use mathematical reasoning tasks for verification, which is a typical problem for complex reasoning tasks.}?}

\begin{figure}
    \includegraphics[width=0.46\textwidth]{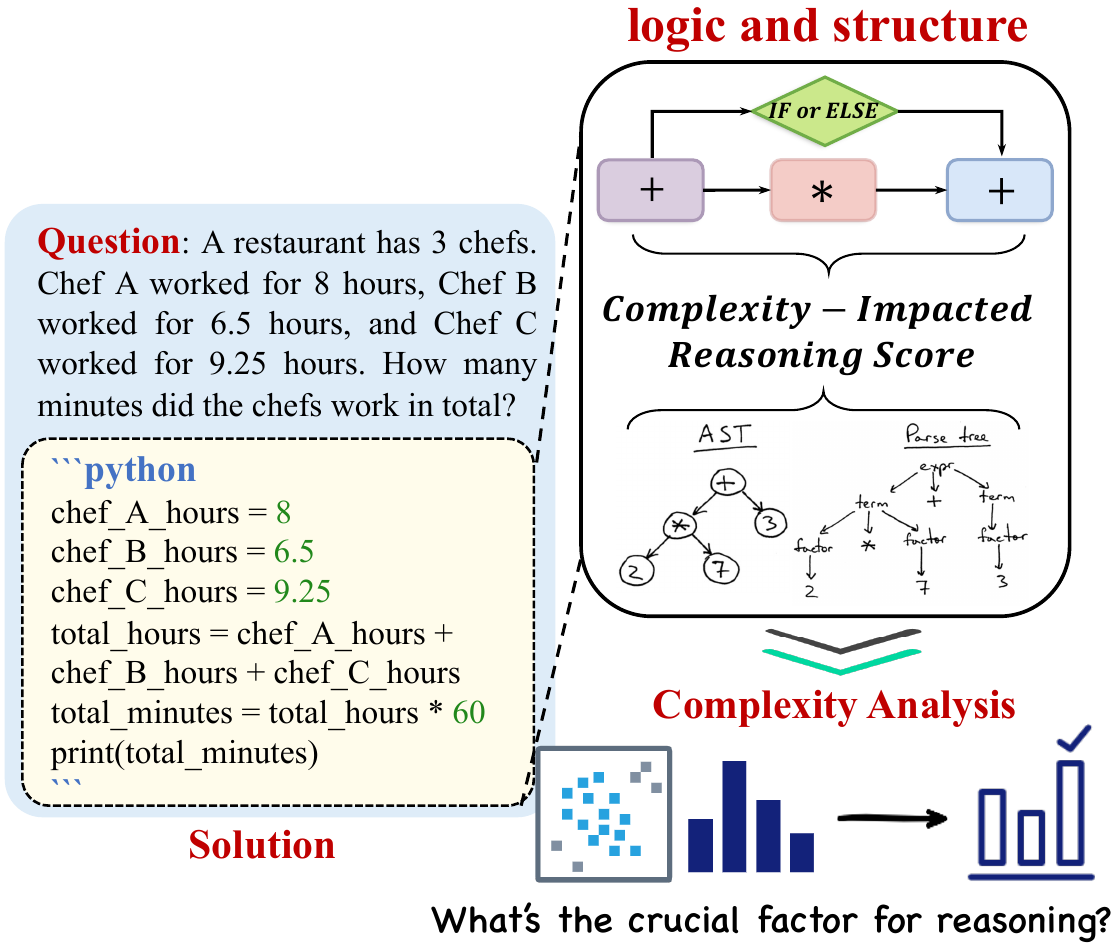}
    \caption{
    We leverage code structure to analyze what kind of data is crucial for reasoning abilities of LLMs models.
   }
    \label{figure:intro}
\vspace{-2mm}
\end{figure}

In this work, we propose the \textbf{\underline{C}}omplexity-\textbf{\underline{I}}mpacted \textbf{\underline{R}}easoning \textbf{\underline{S}}core (\ours), a comprehensive metric for the relationship between code reasoning steps and their impacts on LLMs’ reasoning capacities. 
We postulate that programming languages hold distinct advantages due to:
(1) their superior modeling of intricate structures compared to serialized natural language.
(2) their inherent procedure-oriented logic, which assists in addressing multi-step reasoning problems.
\aaai{We posit that our metric should evaluate the code complexity from both structural and logical perspectives. }

Specifically, we use abstract syntax tree (AST) to calculate the structural complexity of code reasoning steps (rationales). 
To retain all structural information in AST that is represented as a tree, our  approach leverages three AST indicators (node count, node type, depth), which provides a comprehensive understanding of code structures.
Meanwhile, inspired by Halsted \cite{halstead1977elements} and McCabe \cite{mccabe1976complexity}'s theory, we design a method to calculate logical complexity by integrating code difficulty and cyclomatic complexity.
Thus, the operators, operands and control flow of the code can be taken into account. We can explicitly compute the complexity of logical inherent in the code.

Through an empirical analysis by our proposed \ours, we find that not all code data of complexity can be learned and understood by LLMs and
current LLMs have limited understanding of symbolic knowledge like code.
Code blocks with low complexity contain insufficient knowledge, while those with high complexity could be too difficult for LLMs to learn.
Consequently, only code data with an optimal level of complexity (structure\&logic), neither too simple nor too intricate, contribute to the effective enhancement of LLMs' reasoning abilities.

Then, we propose the auto-synthesizing and stratifying algorithm that can automatically generate and \bizhen{filter out the data with the most effective  reasoning ability.}
We apply our algorithm to two scenarios:
(1) guiding instruction generation for mathematical reasoning tasks. 
(2) filtering code data for code generation tasks.
Compared to baseline models, our proposed method achieves favorable results in mathematical reasoning and shows effectiveness for code generation tasks.
In this paper, our contributions are as follows:
\begin{itemize}
    \item We propose a novel method to measure reasoning complexity for the code data, termed {\ours}.
    Our approach, which evaluates the code data from both structural and logical perspectives, can accurately gauges the correlation between code complexity and its reasoning ability.
    
    \item We empirically analyze the impact of varying  complexities, identifying that optimal level of code languages, which is leanable for LLMs, as the pivotal factor in  the reasoning abilities of \textit{program-of-thought}  prompting.

    
    \item We design an auto-synthesizing and stratifying algorithm and apply our approach to both instruction generation for mathematical reasoning and code data filtering for code generation tasks.
    Extensive results demonstrates the validity of our proposed perspective.
\end{itemize}

\section{Background}


Code large language models have demonstrated remarkable capabilities in various tasks such as commonsense reasoning \cite{COCOGEN}, information extraction \cite{Code4Struct}, mathematical reasoning \cite{MathPrompter}, robotics manipulation \cite{DBLP:journals/corr/abs-2307-05973} and embodied learning agent \cite{DBLP:journals/corr/abs-2305-16291}.
Generally, code LLMs with larger model parameters are more effective than vanillar LLMs for reasoning.
We find that even if Codex \cite{Codex} and GPT-3.5 \cite{DBLP:conf/nips/BrownMRSKDNSSAA20} are with same parameters, Codex that is pre-trained on code corpus performs better than GPT-3 on problems such as arithmetic reasoning and structural prediction tasks.
Intriguingly, training on code data not only enables the ability of code understanding but may also foster the reasoning ability.

Inspired by \citet{POT, PAL}, we formalize the multiple-step reasoning tasks by using code-format chain-of-thoughts.
For \textit{program-of-thought} prompting, given the input for the reasoning problem $Q$,  we aim to maximize the likelihood of the answer $A$ as $p(A|Q)$.
\begin{equation}
    p(A|Q) = p(A| {Q}, {R_c}) p({R_c}| {Q})
    \label{equation:2}
\end{equation}
where ${R_c}$ is the solution of the code which will be generated. 
We enhance the effectiveness of solving multi-step reasoning problems by using code prompts as intermediate steps.

\begin{figure*}
    \centering
    \includegraphics[width=0.85\textwidth]{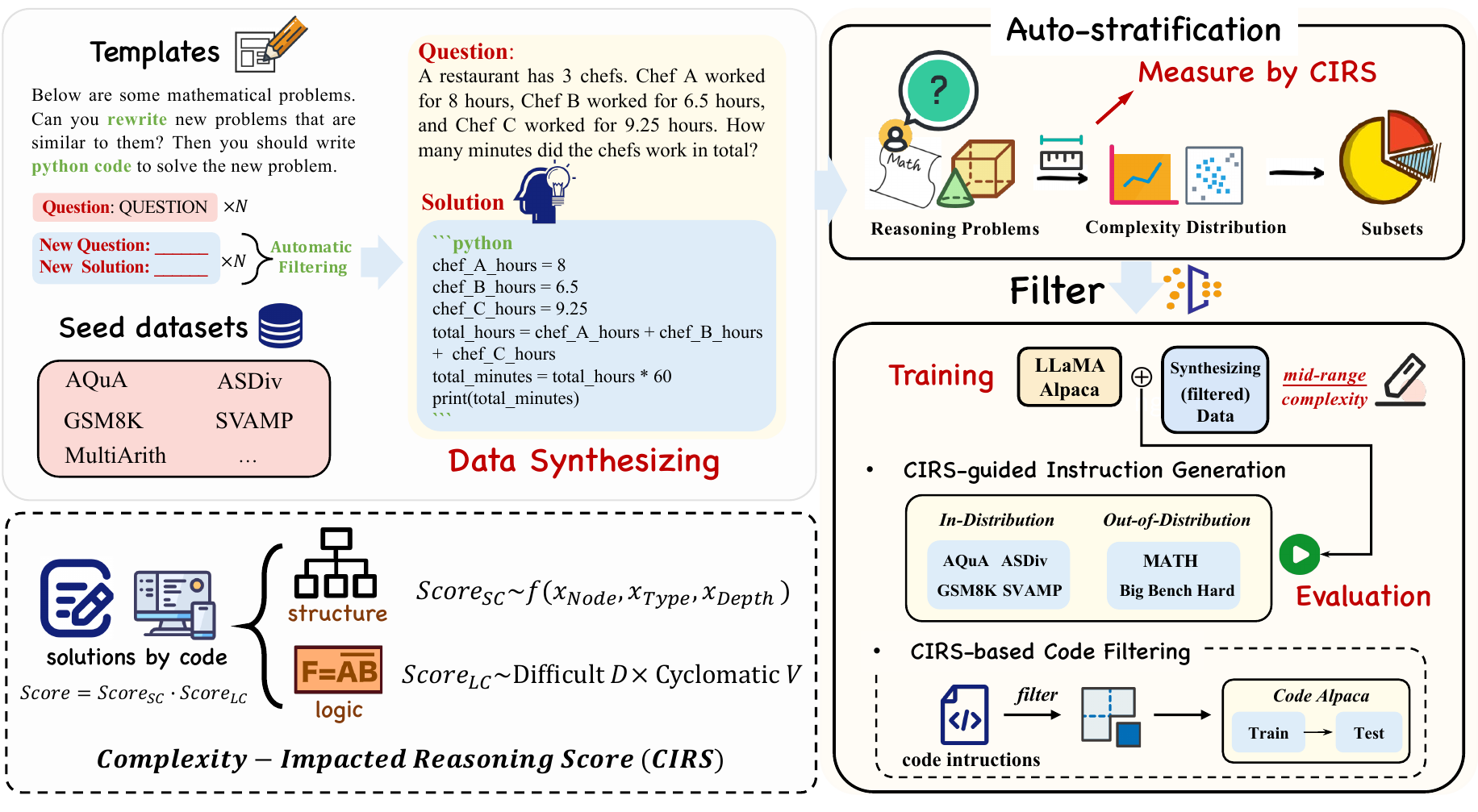}
    \caption{
    We utilize complexity-impacted reasoning score (\ours) to measure the complexity of code reasoning steps. 
    We first synthesize data and employ {\ours} to analyze the complexity distribution of the code reasoning data.
    Then, we analyze and split the data into three different subsets.
    Next, we validate the performance on different model parameters.
    Finally, we leverage the auto-synthesizing and stratifying algorithm and evaluate its performance on the filtered data with the most effective complexity.
    }
    \label{figure:main}
\vspace{-2mm}
\end{figure*}

\section{Complexity-Impacted Reasoning Score}

\label{section:score}


To measure the the reasoning ability of the code rationale $R_c$, we define the complexity-impacted reasoning score as the product of structural complexity 
$\text{Score}_{SC}$ and logical complexity $\text{Score}_{LC}$.
\begin{equation}
{
\text{Score}(R_c)
}
=  \text{Score}_{SC}(R_c) \times \text{Score}_{LC}(R_c)
\end{equation}

\paragraph{Structural Complexity}
To calculate the structural complexity, we measure the structural complexity of the Abstract Syntax Tree (AST). 
We design a simple yet effective method by selecting three indicators that can provide a comprehensive understanding of structural information. 
Therefore, we define the $\text{Score}_{SC}$ as follows:

\begin{equation}
\text{Score}_{SC}(R_c) = \texttt{Sigmoid}(f({{x}}_{\text{Node}},
{{x}}_{\text{Type}},
{{x}}_{\text{Depth}} )) 
\end{equation}

where ${{x}}_{\text{Node}}$, ${{x}}_{\text{Type}}$ and ${{x}}_{\text{Depth}}$ are the features of node count, node types and tree depth in the AST of the code rationale $R_c$.
We first use the function $f$ to apply Z-score normalization to the accumulated data ${x}$ for each feature, and then we aggregate the overall information by \textit{mean pooling}.
Next, we apply the \texttt{Sigmoid} function to transform the data into the range of 0 to 1.
The benefit of doing this is to preserve the distribution characteristics of the feature and avoid being influenced by extreme statistical data, whether it is exceptionally large or small. The detailed explanations for three indicators are as follows:

\begin{itemize}
    \item \textbf{Node Count}. The number of nodes reflects the size of the code. Generally, more nodes indicate higher complexity. But node count alone cannot comprehensively measure code complexity because a large code with a simple structure might be easier to understand than a smaller code with a complex structure.

    \item \textbf{Node Types}. Node types help identify the structural elements present in the code, such as conditional statements, loops, and function calls. Different node types play different roles in the code and contribute differently to its complexity. Therefore, tracking the quantity of various node types can enhance our understanding of the structural complexity of the code.

    \item \textbf{Tree Depth}. The depth of the AST reflects the level of nesting in the code. 
    A greater tree depth may imply more complex control flow and logic, making the code harder to understand. 
    \bizhen{
    It is important that depth alone is also not the sole measurement criterion. 
    A shallow tree with multiple simple branches might be easier to comprehend than a deep tree with a few complex branches.
    }

\end{itemize}

\paragraph{Logical Complexity}
We define code logical complexity $\text{Score}_{LC}$ integrating the difficulty $D$ and cyclomatic complexity $V$, which is inspired  by Halstead Complexity Metrics \cite{halstead1977elements} and McCabe's Cyclomatic Complexity \cite{mccabe1976complexity}.

\begin{equation}
\text{Score}_{LC}(R_c) =
\texttt{Sigmoid} (D(R_c) \times V(R_c))
\end{equation}

where Difficulty $D(R_c) $ denotes the difficulty for solving the problem and $V(R_c) $ means cyclomatic complexity of the rationale $R_c$.
To represent the effort required to comprehend the program, the Difficulty $D(R_c)$ is defined as: 
\begin{equation}
D(R_c) = \left(\frac{n_1}{2}\right) \cdot \left(\frac{N_2}{n_2}\right)
\end{equation}

where $n_1$ denotes the number of distinct operators and
$N_2$ denotes the total number of operands in the code.
$n_2$ denotes the number of distinct operands in the  code rationale $R_c$.
In this formula, the term $(n_1 / 2)$ represents the average complexity of operators, while the term $(N_2 / n_2)$ represents the average complexity of operands. 

To consider the complexity of the logical loops (code control flow), we define the cyclomatic complexity $V(R_c)$ as:

\begin{equation}
V(R_c) = E - N + 2
\end{equation}

where $E$ denotes the number of edges in the control flow graph in the code and $N$ denotes the number of nodes in the control flow graph.
We employ the \texttt{Sigmoid} function to constrain the values of code logical complexity.
There is a significant correlation between potential program errors and high cyclomatic complexity. 
We note that high cyclomatic complexity indicates that the program code has complex judgement logic, potentially leading to lower quality.
It might be difficult to test and maintain those code with  high cyclomatic complexity.
Generally, by integrating the difficulty and cyclomatic complexity, both the complexity of the operators, operands, and control flow of the code can be taken into account. 
Next, we conduct experimental analysis to empirically study the rationality of our method.

\section{Experimental settings}
In order to conduct an unbiased evaluation of all model performances, we use \textit{zero-shot} and \textit{few-shot} settings for evaluation. 
For \textit{zero-shot} setting, we directly presenting mathematical problems to the model for solution generation, without any demonstrations in the input.
For \textit{few-shot} setting, we choose 3-shot for evaluation where we select three in-context examples with rationales.
\aaai{Our criterion for evaluation is that the answer is considered ultimately correct only if the code executor’s answer is correct. }

In {Section \ref{sec:empirical}}, we conduct an empirical analysis of the variations in different model sizes and complexities in \textit{zero-shot} setting.
We construct our own test dataset because there are no publicly available benchmarks up until now.
Model evaluation is performed on AsDiv \cite{ASDiv}, GSM8K \cite{GSM8K}, MultiArith \cite{MultiArith}, and SVAMP \cite{SVAMP}, with a selection of 500 instances randomly chosen from each original testset to form the new testsets. We  chose \textit{gpt-3.5-turbo} as the main benchmark model and accuracy (Acc) as our evaluation metric.

In {Section \ref{sec:results}}, we train the model based on the  LLaMA-7B (Version 1.0) \cite{LLaMA}. 
Vicuna \cite{vicuna2023} and Falcon \cite{falcon40b} are selected as the main comparison models and accuracy (Acc) is chosen as the evaluation metric  again.
Apart from the datasets used in the in-distribution setting, the model's performance is also evaluated on MATH \cite{DBLP:conf/nips/HendrycksBKABTS21} and BigBench-Hard \cite{BBH} in the out-of-distribution setting.
\bizhen{It should be noted that we only choose level-1 problems in MATH.
}
We utilize algorithmic and multi-step arithmetic reasoning tasks in BIG-Bench Hard.
The detailed experimental setup is shown in the supplementary.

\begin{figure*}[h]
\centering
\includegraphics[width=0.80\textwidth]{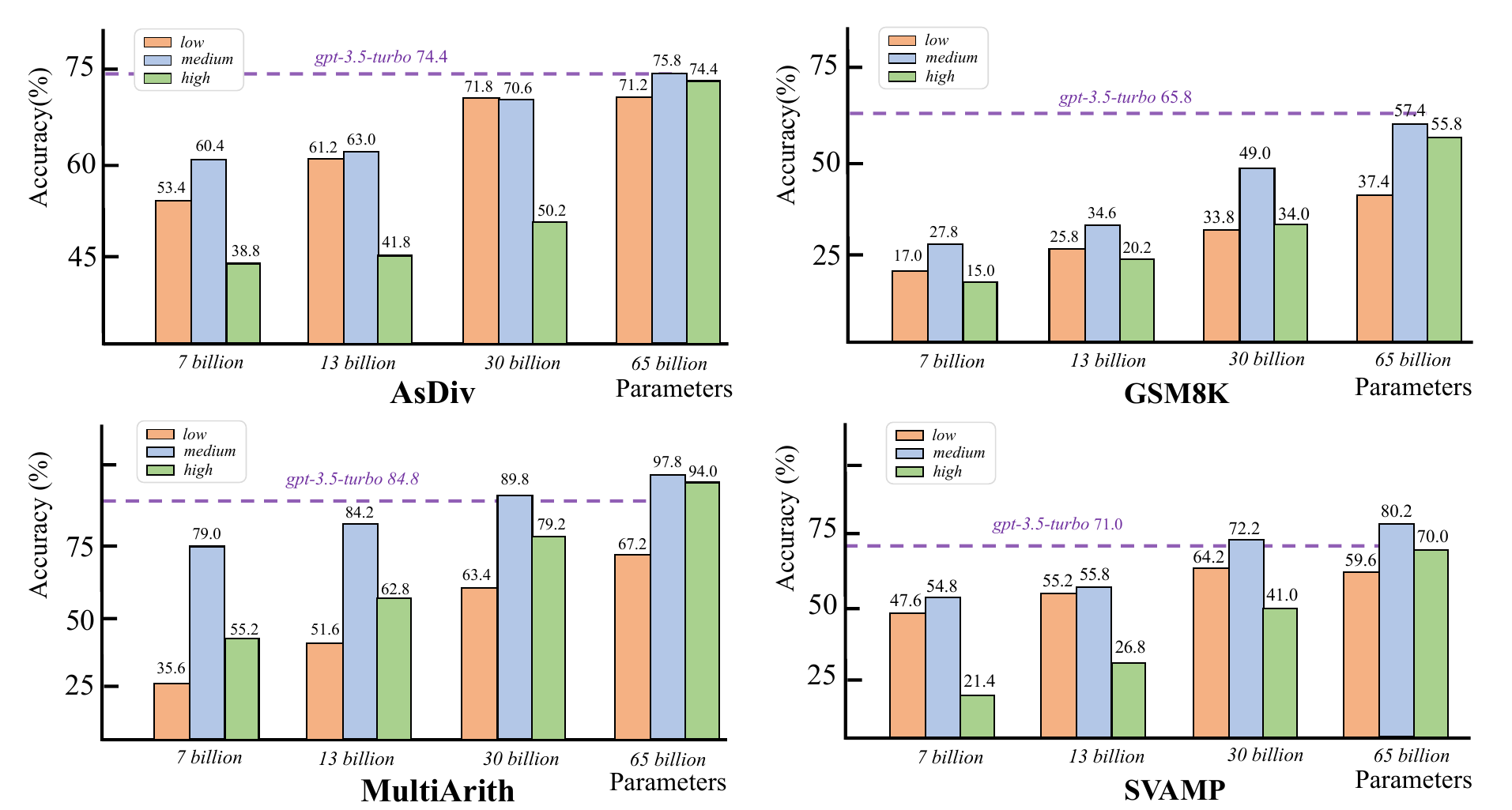}
\caption{
    Evaluation performance on dataset GSM8K, MultiArith, ASDiv and SVAMP.
    We train three models (\textit{low}, \textit{medium}, \textit{high}) whose datasets contain the same number of samples for fair comparison.
    We use Accuracy ($\%$) as the evaluation metrics.
    }
\label{figure:line_chart}
\end{figure*}

\section{Empirical Analysis}
In this section, we empirically analyze the impact of different forms of code data.
Specifically, we synthesize a totally new  dataset and manually partition it using our {\ours} in Section \ref{section:data-synthesizing}.
In Section \ref{section:score-impacts}, we discuss the impact of code data with different complexities on the reasoning abilities for LLMs. 
Then we analyze the characteristics of code data with varying complexities in Section \ref{section:characteristics}.
Finally, we conduct more ablation analysis in Section \ref{section:ablation-distribution}  and \ref{section:ablation-textual}.

\label{sec:empirical}

\subsection{Data synthesizing}
\label{section:data-synthesizing}




\begin{table}[htbp]
    \centering
    \scalebox{0.8}{
    \begin{tabular}{lll}

    \toprule
    \textbf{Seed source} & \textbf{Seed size}  & \textbf{Data size}  \\
    \midrule
     AQuA     & 97,467  & 10,160 \\
     GSM8K   & 7,644  & 12,812 \\
     MultiArith     & 600  & 12,163 \\
     ASDiv      & 2,306  & 13,554   \\
     SVAMP     & 3,954  & 12,441 \\
    \midrule
     ALL  &     & 61,130 \\

    \bottomrule
    \end{tabular}
    }
    \caption{Statistics of seeds and the generated data size.
    }
    \vspace{-2mm}
    \label{table:seed}
\end{table}

To fairly explore the impact of the variations in different complexity scores, it is necessary to avoid errors caused by the dataset itself and generate entirely new forms of code data.
The sources of seed data include the training set of GSM8K \cite{GSM8K}, MultiArith \cite{MultiArith}, Asdiv  \cite{ASDiv}, SVAMP \cite{SVAMP} and AQuA \cite{AQuA}.
In Table \ref{table:seed}, we have synthesized over 60,000 samples from five seed datasets. 
For each dataset, we generate approximately 10,000 samples.
We choose as many datasets as possible to ensure the diversity of mathematical problems.

Then, we design a pipeline that can automatically generate high-quality code corpus by leveraging ChatGPT.
As shown in {Figure \ref{figure:main}}, we apply a template to define the format and then allow the API to continuously rewrite new questions and their corresponding code-format solutions.
In the construction of templates, we randomly select three problems from the seed datasets each time.
\aaai{Next, we automatically filter out the generations that do not conform to Python syntax standards, which results in a collection of high-quality mathematical problems.
For all generated data, we randomly sampled\ 10\% and verified its correctness
by manual checks and automated validation with GPT-4, ensuring the accuracy within a reasonable margin of error.
}

After obtaining well-generated code data, we utilize {\ours} ({Section \ref{section:score}}) and manually split the data into different subsets based on the analysis of code complexity distribution.
We put the visualized results in the supplement.
Based on different complexity scores, we  name the partitioned subsets as \textit{low} (lower score samples), \textit{medium} (medium score samples) and \textit{high} (high score samples).

\begin{figure*}[t]
    \centering
    \includegraphics[width=0.75\textwidth]{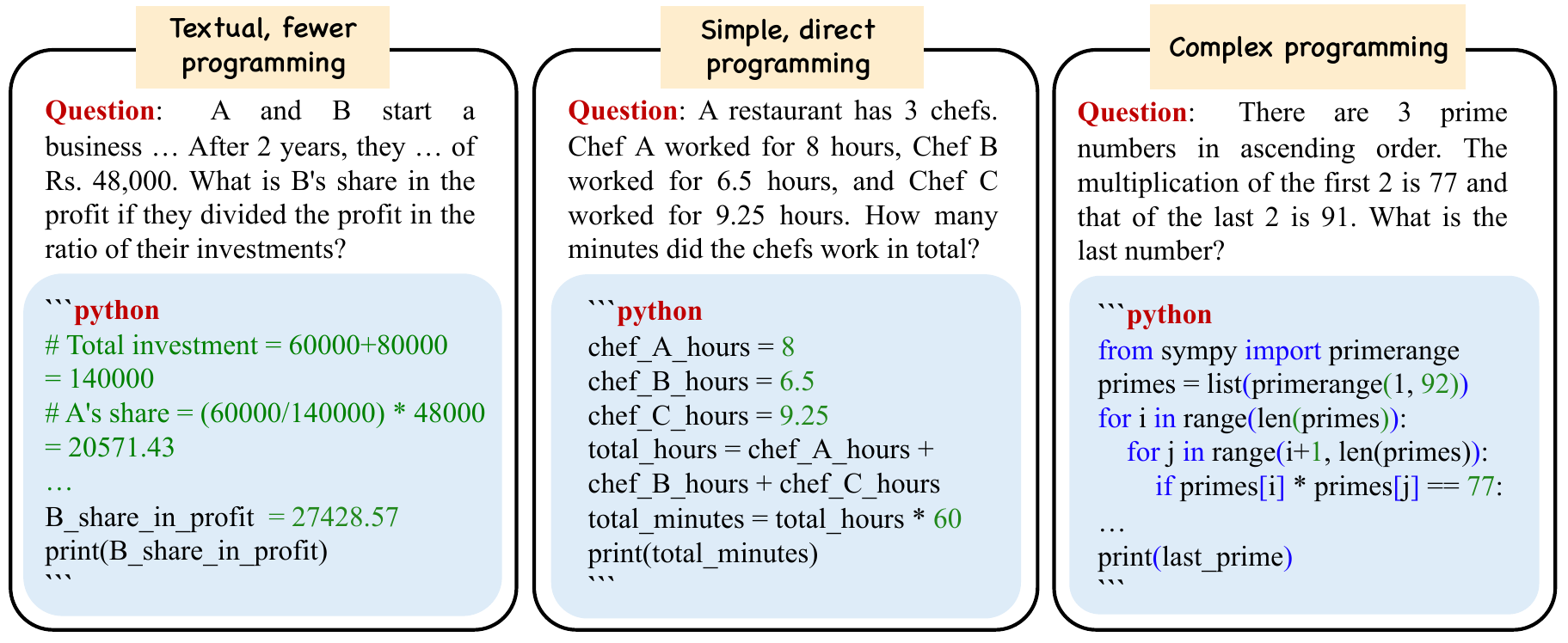}
    \caption{
         {
         {
         As the {\ours} score increases, there is a greater presence of logical and structural information in the code.        
         }
         }
        }
    \label{figure:example-score}
\end{figure*}

\subsection{Impacts of different complexity 
score}
\label{section:score-impacts}

To compare the impact of different code complexities on the reasoning capability of LLMs,
we train three models based on LLaMA (Version 1.0) from 7 billion to 65 billion parameters.
We randomly select 1,700 instances from each subset (\textit{low}, \textit{medium}, \textit{high})  to build the training and validation dataset for fair comparisons. 
Results are shown in Figure \ref{figure:line_chart}.

\textbf{ (1) Optimal level of code is crucial to the reasoning abilities of \textit{program-of-thought} prompting.
}
From the results across the four datasets, we note that the model performs optimally when the complexity of the code data is in mid-range. 
This suggests that the learnable symbolic language is crucial to the reasoning abilities of program-aided prompting.
The reasoning behind this is that data with overly simplistic complexity, is too simple for LLMs, leading to less noticeable effects. 
Conversely, when the complexity escalates significantly, the logical semantics and nested structures become difficult to comprehend or {learn}, which could adversely impact the reasoning capabilities of LLMs.

\textbf{(2) The larger the number of parameters, the more significant the gain in LLM's reasoning capabilities.}
It is evident that as the model size increases from 7 billion to 65 billion , the effectiveness of its reasoning capability improves. In fact, after fine-tuning, most 65 billion parameter models can achieve results comparable to those of \textit{gpt-3.5-turbo}. 
\bizhen{It suggests that having a sufficient number of parameters is crucial for  substantial reasoning capabilities in language models. }
Furthermore, when the language model is large enough, the difference in results across various complexities is minimal. 
This indicates that LLMs with vast parameters are more prone to symbolic data and  inherently have the potential to yield strong reasoning capabilities.

\textbf{ (3) Current LLMs have limitations in their understanding capabilities for reasoning.}
We observe that when data complexity is extremely high, the performance of LLMs tends to decrease. It reflects that there is an inherent limit to the reasoning capabilities of large language models.
We argue that: 
(1) The current architecture of LLMs (such as \textit{decoder-only LLM}) has limited ability to understand complex knowledge, which also restricts the emergence of their reasoning capabilities. The prerequisite for large models to demonstrate powerful reasoning abilities is their ability to comprehend the structures and logical knowledge embedded in complex data.
Therefore, it is necessary to explore model structures with stronger reasoning abilities in future research.
(2) Further enhancement in reasoning power requires the reliance on external tools.
We know that the scope of reasoning problems is quite broad, not only mathematical reasoning, but also including commonsense or more complex logical reasoning tasks.
Therefore, relying solely on the LLM itself is not enough to resolve all issues at once; the assistance of more powerful external tools is required.

\subsection{The characteristics of different {\ours} scores.} 
\label{section:characteristics}

In Figure \ref{figure:example-score}, we investigate the characteristics of different {\ours} scores.
The different subsets of {\ours} scores exhibit distinct structural and logical differences.
Inspired by \cite{haladyna1997writing, conklin2005taxonomy}
and AoPS\footnote{\url{https://artofproblemsolving.com/}}, 
we also find the results of different complexity scores correspond to the cognitive level of difficulty for reasoning problems. 


\begin{itemize}
    \item \textbf{Textual, minimal programming}.
    Samples with lower {\ours} scores contain little structural information.
    Although they do contain some intermediary reasoning processes, these are primarily represented in flat textual descriptions. 
    These samples typically correspond to  simpler and   structurally, logical insufficient problems.
    
    \item \textbf{Simple but direct programming}.
    As {\ours} score increases in the code reasoning steps, the presence of programming languages with simple logical semantics and structures also escalates.
    These samples typically involve simple and straightforward logical operations. 

    \item \textbf{Complex programming}.
    Samples with exceedingly high scores contain substantial amounts of structural function definitions or reasoning processes, which suggests the presence of numerous complex conditional statements and function structures.
    These samples are typically highly challenging mathematical problems.

\end{itemize}

\subsection{Excluding the effect of the complexity distribution itself}

\label{section:ablation-distribution}

   \begin{figure}[]
        \centering
        \includegraphics[width=0.45\textwidth]{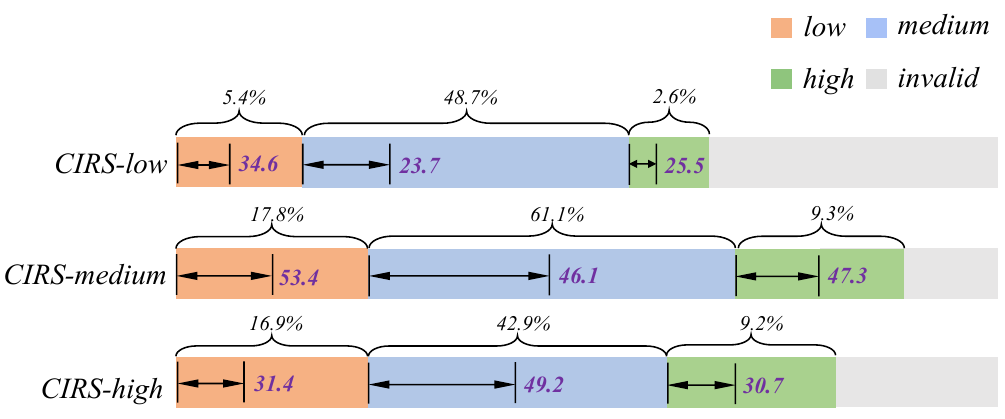}
        \caption{
        {
        Ablation analysis for different code complexities.
        We use {\ours} to measure the  predictions for each model and divide them into four categories 
        (\textit{low}, \textit{medium}, \textit{high} and \textit{invalid}).
        $\overbrace{}$ means the percentage of output predictions and $\longleftrightarrow$ denotes the prediction result of each category (accuracy $\%$).
        The results show that
        the effectiveness of complexity data is not
        because of the frequency of data occurrence.
        }
    }
    \label{figure:excluding}
    \end{figure}

To negate the potential skew from data distribution itself, such as enhanced performance in the mid-range data due to its higher frequency of occurrence, we conduct a more in-depth analysis of the {evaluation} results at different complexity scores. 
We use the trained 7B model in Section \ref{section:score-impacts} and conduct tests on 2,000 samples with three models 
(\textit{CIRS-low}, \textit{CIRS-medium}, \textit{CIRS-high}).
It should be noted that we use {\ours} to measure the output reasoning steps for each model and divide them into four categories (\textit{low}, \textit{medium}, \textit{high} and \textit{invalid}).
From the results in Figure \ref{figure:excluding},  we find that \textit{CIRS-medium} generates the highest number of valid predicted outputs in three distributions ($ \textit{17.8\%}$, $\textit{61.1\%}$, $\textit{9.3\%}$).
We also observe that \textit{CIRS-medium}  demonstrates high accuracy ($\textit{53.4}$, $\textit{46.1}$, $\textit{47.3}$) in all three  distributions. 
The accuracy of predictions for each distribution by the model is independent of the quantity of training data.
Therefore, we can conclude that the effectiveness of complexity data is not because of the frequency of data occurrence.

\subsection{Ablation analysis for textual rationales}
\label{section:ablation-textual}

\begin{figure}
\centering
\includegraphics[width=0.38\textwidth]{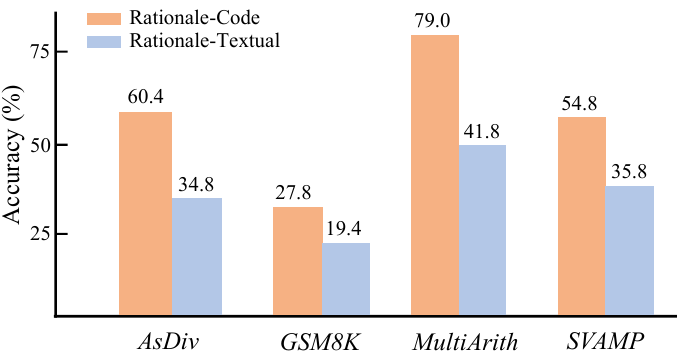}
\caption{
{Comparison for textual and code rationales.
We use Accuracy($\%$) as the evaluation metrics. 
Training with code data demonstrates a clear advantage in all datasets.
}
}
\label{figure:ablation-textual}
\end{figure}

\bizhen{
To verify the effect of code and textual rationales,
we substitute the code-format solving process with textual rationales using the same datasets.
} 
We sample 1,700 instances of code data within the mid-range complexity and simultaneously construct a dataset that uses textual rationales. 
We train both two models based on LLaMA-7B.
As shown in Figure \ref{figure:ablation-textual}, the code dataset demonstrates a clear advantage in all four datasets.
It is because code inherently encapsulates logical semantics and structural information.
Another reason is that code can be executed by external interpreters.
So solutions with code are superior to flattened textual information.

\section{CIRS for Improving the Reasoning Ability}
\label{sec:results}

In this section, we describe our auto-synthesizing and stratifying algorithm in Section \ref{sec:results-algorithm}.
Then we apply {\ours} to instruction generation for mathematical reasoning, code data filtering for code generation tasks in Section \ref{sec:results-usage1} and \ref{sec:results-usage2}.

\subsection{Auto-Synthesizing and Stratifying}
\label{sec:results-algorithm}
Based on the processing step in Section \ref{sec:empirical}, we formalize the whole procedure into a pipeline method for automatic data generation and stratification.
The auto-synthesizing and stratifying algorithm is described in Algorithm \ref{alg:auto-stratify}.

We first do the template $T$ filling by calling APIs and get the \bizhen{synthesized} dataset $D$.
Then we calculate the distribution of complexity for all synthesized data by {\ours} and get the threshold set $J$.
Next we design a threshold-based $k$-means clustering method that automatically  partitions the dataset according to complexity characteristics. 
Finally, we will apply our proposed algorithm for two scenarios to enhance the reasoning abilities of LLMs.

\begin{algorithm}[]
    \small
    \caption{Auto-Synthesizing and Stratifying}
    \label{alg:auto-stratify}
    \begin{algorithmic}[1]
    \REQUIRE $T$: Template, $K$: Number of clusters, $J$: Threshold set
    \ENSURE $C$: Cluster assignments
    \STATE Dataset $D \gets$ template $T$ filling by leveraging API
    \STATE Threshold $J \gets$ threshold set generated by {\ours}
    \STATE Initialize $C$ with random initial cluster assignments
    \REPEAT
        \STATE Clear all clusters
        \FOR{each data point $x$ in $D$}
            \STATE Find the nearest centroid $c_i$ in $C$ to $x$
            \STATE Assign $x$ to cluster $c_i$
        \ENDFOR
        \FOR{each cluster $c_i$ in $C$}
            \STATE Recalculate centroid $c_i$ as the mean of all points assigned to $c_i$
        \ENDFOR
        \STATE Remove clusters from $C$ if the average distance to their centroid is not in $J$
    \UNTIL{no more updates or maximum iterations reached}
    \RETURN $C$
    \end{algorithmic}
\end{algorithm}

\subsection{Usage1: CIRS-guided Instruction Generation}
\label{sec:results-usage1}

\begin{table*}[t]
    \centering
    \renewcommand{\arraystretch}{1.2}
    \scalebox{0.8}{
    \begin{tabular}{lccccccc}
    \toprule[1.25pt]
    \multirow{3}{*}{\textbf{Models}} &  & \multicolumn{6}{c}{\textbf{Mathematical Reasoning}}  \\
    \cline{3-8} 
   & \textbf{Parameters} & \multicolumn{4}{c}{\textbf{In-Distribution}} & \multicolumn{2}{c}{\textbf{Out-of-Distribution}}   \\

    \cmidrule(lr){3-6}
    \cmidrule(lr){7-8}
    &  & \textbf{AsDiv}  & \textbf{GSM8K}  &  \textbf{MultiArith} &  \textbf{SVAMP} &  \textbf{MATH} & \textbf{BigBench-Hard$\dagger$}  \\
    
    \midrule[1.1pt]
    \rowcolor[rgb]{0.93,0.93,0.93}
    \multicolumn{8}{l}{\textit{Zero-shot, Answer-only Prompting}} \\

    {\textbf{Falcon*}} & 7B & 14.7  & 3.6 & 6.0 & 5.6  &  4.8 & 19.0 \\
   {\textbf{Vicuna}} & 7B & 35.8  & 8.6  & 16.4  & 33.0  &  14.5 &  25.3   \\

    \textbf{gpt-3.5-turbo } &  \textbf{/} & {74.4}   & {65.8} & {84.8} & {71.0} & {70.1} & {37.3} \\
    \hline
    {\ours} \textbf{(LLaMA)} & 7B & {69.2} & {40.4} &  {97.2}  & {70.2}  & {38.6} &  {37.7} \\

    \midrule[1.1pt]
    \rowcolor[rgb]{0.93,0.93,0.93}
    \multicolumn{8}{l}{\textit{Few-shot, Chain-of-thought Prompting}} \\
    \textbf{Falcon} & 7B & 7.9 & 3.0 & 5.4 &4.6 & 2.9& 23.2 \\
   {\textbf{Vicuna}}  & 7B  &  34.9 & 9.1  &  17.2 & 32.0  & 17.2  &  35.3 \\
    \textbf{gpt-3.5-turbo } &  $ \textbf{/} $  &  80.6  & 61.4 & 44.8 & 71.6 & 68.5 & 50.1\\
    \midrule
    {\ours} \textbf{(LLaMA)} & 7B & 65.4&37.6 & 96.0& 69.4 & 39.2 & 36.3 \\

    \bottomrule[1.25pt]
    \end{tabular}
    }
    \caption{
        Results of mathematical reasoning tasks. $\dagger$ We choose algorithmic and multi-step arithmetic reasoning tasks in BIG-Bench Hard.
        *Here we use \textit{Falcon-Instruct} which is fine-tuned on instruction datasets. 
    }
    \label{table:main}
    \end{table*}

From the analysis in {Section \ref{sec:empirical}}, 
\bizhen{
we know that the trained model with 
complexity optimal level of code data, exhibits the best reasoning capabilities.
Therefore, we employ our Algorithm \ref{alg:auto-stratify} to filter out more data from the source dataset to train an enhanced reasoning model, specifically targeting the mid-range complexity range.
}
Totally, we collect 40,000 data samples to train a more powerful language model for reasoning.
Results are shown in Table \ref{table:main}.
For in-distribution setting, we find that trained model outperforms Vicuna and Falcon.
To eliminate the influence of data distribution, we directly test the model's performance in the out-of-distribution setting.
Our model perform best (the same parameters) in both \textit{zero-shot} and \textit{few-shot} prompting.
{It is worth noting that our approach  demonstrates comparable effectiveness to ChatGPT on BigBench-Hard in \textit{zero-shot setting}.}
For MATH dataset, we notice that our model still outperforms the baseline models. 
But our model are much worse than ChatGPT which is due to  limitation of code data itself.

\subsection{Usage2: {\ours}-based Code Filtering}
\label{sec:results-usage2}

\begin{table}[ht]
    \centering
    \scalebox{0.8}{
    \begin{tabular}{lll}
    \toprule
    
    \textbf{Models} & \textbf{Parameters} & \textbf{Acc.}   \\
    \midrule
    \textbf{Alpaca}   & 7B &   24.0   \\
    \midrule
    \textbf{Code-LLaMA}  & 7B &  50.0 \\
   \textbf{Code} ({\ours})\textbf{-LLaMA}  & 7B &  \textbf{55.0} \\
    \bottomrule
    \end{tabular}
    }
    \caption{
    Results of CIRS-based code filtering tasks.
    }
    \vspace{-2mm}
    \label{table:code}
\end{table}

To validate the effectiveness of our approach in code-related tasks, we use the Algorithm \ref{alg:auto-stratify} to filter a batch of code instruction data.
We first split the Code Alpaca \cite{codealpaca} into train and test dataset.
We leverage the whole train dataset to train LLaMA-7B and the trained model is \textit{Code-LLaMA}.
For fair comparison, we  filter the train dataset and get the subset with much more high-quality code instructions.
We train \textit{Code (CIRS)-LLaMA} based on the filtered data.
The results illustrate that \textit{Code (CIRS)-LLaMA} demonstrates effective performance in pure code generation tasks. 
We can conclude that the optimized  structures and logical semantics is most beneficial for LLM's reasoning abilities.

\section{Related Work}


\paragraph{Program-aided Prompting}
Program-of-thoughts \cite{POT} prompting delegates computation steps to an external language interpreter and \cite{PAL} generates programs as the intermediate reasoning steps.
\cite{binder} is a  neural-symbolic framework that maps the task input to a program.
Similarly, \cite{DBLP:journals/corr/abs-2305-18507} is  a neural symbolic prompting method for complex reasoning tasks.
Some methods such as \cite{Code4Struct, codeie, DBLP:journals/corr/abs-2304-09048} leverages code prompting methods for information extraction tasks.
\citet{COCOGEN} frames the task of structured commonsense reasoning as code generation.
\cite{DBLP:journals/corr/abs-2305-13888}  distills LLMs into specialized, compact models for reasoning tasks by program-aided prompting.



\paragraph{Reasoning with Large Language Models}
The research on reasoning abilities is a core issue in NLP \cite{ACL2023_PromptReasoningSurvey, DBLP:journals/corr/abs-2212-10403, DBLP:journals/corr/abs-2303-18223}.
The success of LLMs have progressively achieved a series breakthroughs in various  tasks or domains \citep{MathPrompter, LogicSolver, DBLP:journals/corr/abs-2212-08686, chen2023teleknowledge}.
\aaai{
Some research studies \cite{DBLP:journals/corr/abs-2305-19555, DBLP:journals/corr/abs-2305-19213, DBLP:journals/corr/abs-2305-12096,yuan2023scaling, schwartz2020right} are focusing on analyzing the  capabilities of large models themselves.
}
\cite{wang2023making} improves LLMs reasoning abilities by fine-tuning alignment paradigm.
More and more research efforts \cite{ DBLP:journals/corr/abs-2301-12726, DBLP:journals/corr/abs-2306-02707} are being devoted to unveiling the origin of a model's reasoning abilities or focus on enhancing the capability of smaller models. 
Some works  \cite{DBLP:conf/emnlp/WiegreffeMS21, DBLP:journals/corr/abs-2305-00633}
generate rationales to enhance model interpretability.
To measure reasoning capabilities, \cite{DBLP:conf/iclr/FuPSCK23} propose a selection scheme based on complexity prompting.
\cite{DBLP:journals/corr/abs-2305-17306} is an open-source evaluation suite that measures LLMs' multi-step reasoning performance.
Different from previous work, our work is the first to analyze the reasoning capabilities of large language models from code data.


\section{Discussion and  Conclusion}

\textit{What kind of data format is crucial for LLM's reasoning abilities?}
We explore the reasoning abilities for \textit{program-of-thought} prompting and
the results indicate that code data with optimal level of code, characterized by certain logical and structural qualities, is the key factor. 
Code data is efficient because it is inherently semi-structured and abundant in the natural world. 
We can prove that: (1) The local structural properties of the data are crucial for improving reasoning abilities, which aligns with \cite{DBLP:journals/corr/abs-2304-03843}. 
The logical coherence or a certain amount of knowledge circuitry inherent in the data is necessary. 
(2) Overly complex structural information and logic are `too difficult to learn' for LLMs. 
The experimental results of this work demonstrate that knowledge of optimal level complexity is most effective because it is learnable for most large language models. 
Meanwhile, we also find that as the number of parameters in language models increases, their understanding of complex knowledge also improves.

In this work, we introduce {\ours} to measure the relation between code reasoning steps and reasoning abilities. 
By considering both structural and logical attributes of code data, we use AST to encode the structural information and encode structural feature by difficulty and cyclomatic complexity.
Through an empirical  analysis, we find that 
optimal level of code languages
plays a crucial role in the reasoning abilities of \textit{program-of-thought}  prompting.
We develop the auto-synthesizing and stratifying algorithm that applies mathematical reasoning and code generation tasks.
Extensive results prove the effectiveness of the proposed method. 
\bizhen{
In the future, we will expand this work to more scenarios such as commonsense or logical reasoning tasks and train powerful reasoning models with low computational cost.
}

\section*{{{Acknowledgements}}}
We would like to express gratitude to the anonymous reviewers for their kind comments. This work was supported by the National Natural Science Foundation of China (No. 62206246), the Fundamental Research Funds for the Central Universities (226-2023-00138), Zhejiang Provincial Natural Science Foundation of China (No. LGG22F030011), Ningbo Natural Science Foundation (2021J190), Yongjiang Talent Introduction Programme (2021A-156-G), CCF-Baidu Open Fund, and Information Technology Center and State Key Lab of CAD\&CG, Zhejiang University, and NUS-NCS Joint Laboratory (A-0008542-00-00).

\bibliography{aaai24}
\end{document}